%% file: main.tex
\documentclass{article} 

\usepackage[accepted]{icml2024}

\usepackage{hyperref}
\usepackage{url}
\usepackage{subfigure}
\usepackage{enumitem}
\newenvironment{itemize*}%
 {\leftmargini=20pt\begin{itemize}%
  \setlength{\itemsep}{3pt}%
  \setlength{\parskip}{0pt}%
  }%
 {\end{itemize}}
\newenvironment{enumerate*}%
 {\begin{enumerate}%
  \setlength{\itemsep}{0pt}%
  \setlength{\parskip}{0pt}}%
 {\end{enumerate}}

\usepackage{graphicx}
\usepackage{listings}
\usepackage{titlesec}
\usepackage{booktabs}       
\usepackage{multirow}
\usepackage{makecell}
\usepackage[capitalize,noabbrev]{cleveref}

\mathchardef\mhyphen="2D

\icmltitlerunning{Preprint}

\begin{document}
\twocolumn[
\icmltitle{Investigate-Consolidate-Exploit: \\
A General Strategy for Inter-Task Agent Self-Evolution}

\icmlsetsymbol{equal}{*}

\begin{icmlauthorlist}
\icmlauthor{Cheng Qian}{equal,THU}
\icmlauthor{Shihao Liang}{equal,HKU}
\icmlauthor{Yujia Qin}{THU}
\icmlauthor{Yining Ye}{THU}
\icmlauthor{Xin Cong}{THU}
\icmlauthor{Yankai Lin}{RUC}
\icmlauthor{Yesai Wu}{ModelBest}
\icmlauthor{Zhiyuan Liu}{THU}
\icmlauthor{Maosong Sun}{THU}
\end{icmlauthorlist}

\icmlaffiliation{THU}{Tsinghua University}
\icmlaffiliation{HKU}{The University of Hong Kong}
\icmlaffiliation{RUC}{Renmin University of China}
\icmlaffiliation{ModelBest}{ModelBest Inc.}

\icmlcorrespondingauthor{Yankai Lin}{yankailin@ruc.edu.cn}
\icmlcorrespondingauthor{Zhiyuan Liu}{liuzy@tsinghua.edu.cn}

\icmlkeywords{Machine Learning, ICML}

\vskip 0.3in
]

\printAffiliationsAndNotice{\icmlEqualContribution} 

\input{sections/0_abstract}

\input{sections/1_introduction}

\input{sections/3_preliminary}
\input{sections/3_method}
\input{sections/4_experiment}
\input{sections/5_discussion}

\input{sections/2_related_work}
\input{sections/6_conclusion}

\bibliography{icml2024}
\bibliographystyle{icml2024}

\input{sections/Appendix}

\end{document}

%% file: sections/0_abstract.tex
\begin{abstract}
This paper introduces Investigate-Consolidate-Exploit (ICE), a novel strategy for enhancing the adaptability and flexibility of AI agents through inter-task self-evolution. Unlike existing methods focused on intra-task learning, ICE promotes the transfer of knowledge between tasks for genuine self-evolution, similar to human experience learning. The strategy dynamically investigates planning and execution trajectories, consolidates them into simplified workflows and pipelines, and exploits them for improved task execution. Our experiments on the XAgent framework demonstrate ICE's effectiveness, reducing API calls by as much as 80\% and significantly decreasing the demand for the model's capability. Specifically, when combined with \texttt{GPT-3.5}, ICE's performance matches that of raw \texttt{GPT-4} across various agent tasks. We argue that this self-evolution approach represents a paradigm shift in agent design, contributing to a more robust AI community and ecosystem, and moving a step closer to full autonomy.
\end{abstract}

%% file: sections/1_introduction.tex
\section{Introduction}


\begin{figure*}[!t]
    \centering
    \subfigure{\includegraphics[width=0.8\textwidth]{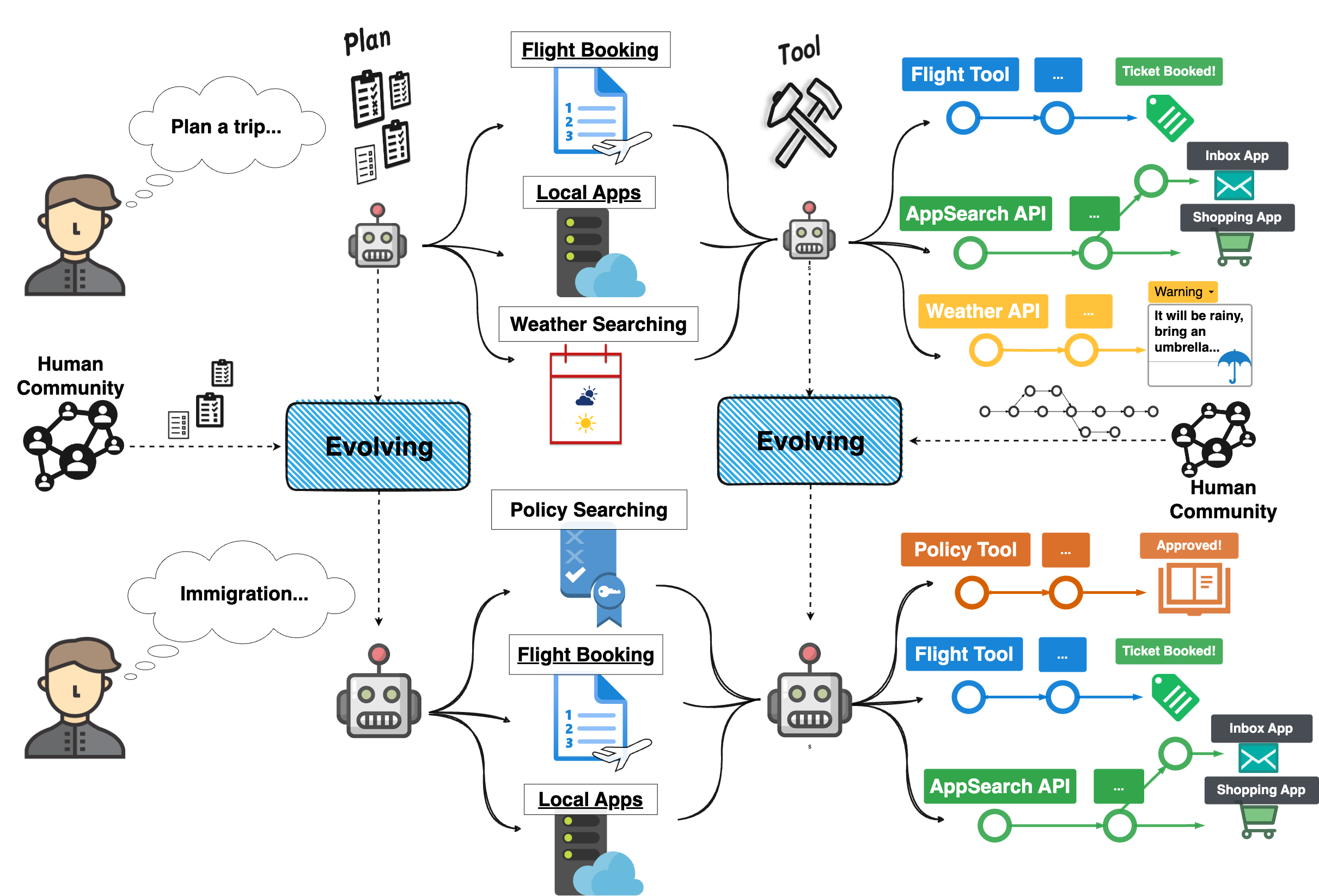}}
    \caption{\small{An overview of the inter-task agent self-evolution. ICE automatically identifies re-utilizable plans and tool execution trajectories as past experiences for agent self-evolution. The human effort may also be involved in crafting experiences for learning.}}
    \label{fig:intro}
\end{figure*}

\label{sec:intro}

The trajectory of human evolution unfolds as a continual process of assimilating and distilling experiences from the past, perpetually advancing the frontiers of human capabilities. This self-directed evolution involves incorporating insights from past failures within tasks to rectify errors, and drawing upon successful experiences across tasks to enhance effectiveness and efficiency. Current language models including the GPT \citep{openai2022chatgpt, openai2023gpt4} and LLaMA \citep{touvron2023llama, touvron2023llama2} series have showcased a remarkable capacity to engage in sophisticated task-solving as agents. While they can leverage tools to address specific capability-related challenges within the task, these agents inherently lack the innate capacity to glean insights from past successes and failures to self-evolve.


Enabling language model-driven agents to assimilate prior experiences, akin to human-like experiential learning, has inspired a spectrum of approaches. These include the integration of post-failure reflections into the model's context \citep{shinn2023reflexion, miao2023selfcheck, qian2023creator}, adaptive optimization of prompts for the models to accommodate instructions \citep{zhou2022large, yang2023large, pryzant2023automatic, hsieh2023automatic}, and retrieval of past contexts for more coherent and effective generation \citep{zhao2023expel, hu2023chatdb, liu2023thinkinmemory, zhong2023memorybank}. These investigations primarily focus on \emph{intra-task} learning, addressing the execution of a specific task. However, they diverge from authentic self-evolution scenarios, which prioritize the transfer of \emph{inter-task} experiences.



Given humans cultivate cognitive flexibility and problem-solving skills through diverse inter-task experiences, it is similarly significant for agents to generalize past knowledge to tackle new challenges. This contributes to \textbf{inter-task agent self-evolution}, which enables the agent's autonomous adaptation and improvement of performance over time. In inter-task scenarios, referencing the previous execution outcomes or directly replicating the entire previous execution process, as seen in intra-task learning methods, proves impractical. This underscores the necessity of disentanglement in experience for effective re-utilization of past experiences.

In the context of current agent designs, a complete experience usually involves two aspects: i) \emph{Planning}, which entails understanding intricate user task objectives and decomposing them into manageable units; ii) \emph{Execution}, which entails a sequence of interactions with the environment through tool invocation and feedback processing.
The blending of both experiences usually hampers the assimilation of pertinent knowledge, impeding adaptability in tackling diverse challenges. Therefore, advocating for their separation within the agent's self-evolution strategy is crucial to facilitate experience learning and re-utilization.
In \Cref{fig:intro}, we illustrate how plan and execution trajectories as past experiences could respectively be applied for agent self-evolution.
Despite this disentanglement, challenges persist in what contents are worth recording as experience, how to standardize their formats for convenient inter-task learning, and when to apply them for future use to raise task effectiveness and efficiency.

In this work, we propose \textsc{Investigate-Consolidate-Exploit} (ICE), the first strategy that enables inter-task self-evolution of general agent designs. ICE disentangles task planning records and execution trajectories as the past experiences respectively for inter-task knowledge transfer, thus promoting both efficiency and effectiveness when handling new tasks. Specifically, ICE is divided into three stages: (1) \textbf{Investigate}: To identify experiences that are worth learning and referencing, we track the plan and status of each decomposed goal and extract successful execution trajectories for all the goals being handled.
(2) \textbf{Consolidate}: To standardize the format of experiences to make future re-utilization automatic and convenient, we prune the plan into a linearized flow of successfully achieved goals (\emph{workflow}), transform the mined trajectories into finite automata (\emph{pipeline}) that enable automated execution for specific purposes, and store them in the database as agent's memory.
(3) \textbf{Exploit}: To enhance the efficiency and effectiveness of new tasks by promptly accessing and utilizing previously consolidated experiences, we retrieve the consolidated plans as in-context references for generating and refining new plans, and directly apply the consolidated trajectories with similar goals for automated execution.


To validate the effectiveness of our approach, we conducted a series of experiments utilizing the XAgent framework \citep{xagent2023} for its clear disentanglement of agent planning and execution, which provides an ideal testbed for validating two self-evolution aspects. Through case studies and controlled experiments, we demonstrate that ICE can i) reduce model API calls by up to 80\%, which significantly saves computational resources; ii) diminish the demands on models' intrinsic abilities, which lowers the barrier for agent deployment. Specifically, when paired with \texttt{GPT-3.5}, our ICE strategy can rival the performance of \texttt{GPT-4} across diverse agent tasks.

Overall, the ICE strategy not only makes agent task completion more effective but also renders its execution more time-efficient and grounding more cost-efficient. All these aspects reflect the success of agent self-evolution.


%% file: sections/3_preliminary.tex
\section{Preliminaries}

The experience accumulated during agent task handling usually involves planning and execution. The planning process is crucial in understanding user intentions and specifying detailed subgoals, while the execution process entails interaction with the environment to implement the goals in the plan. In general, we disentangle the agent's planning and execution experiences to facilitate their re-utilization.


For \textbf{planning}, we aim for a finer granularity in dividing user goals. The utilization of a tree structure serves as a typical and effective method for embodying this hierarchical representation. In the plan tree, the tree root represents the user's ultimate goal $G$, and each non-root node acts as a subgoal under its parent. For instance, the ultimate user goal $G$ could be broken down into $m$ subgoals $G_1, \ldots, G_m$, while $G_1$ could be further broken down into $n$ subgoals $G_{1\mhyphen 1}, \ldots, G_{1\mhyphen n}$. Each subgoal $G_x$ (where $x = x_1\mhyphen x_2\mhyphen \dots$ thus representing any subgoal) may encapsulate meta-information such as milestones that ought to be achieved, suggestions on what to follow or avoid, etc.
In this way, the whole plan tree, along with all its meta-information, serves as a holistic planning experience, instructing the agent in rational and effective decomposition of user goals. In addition, the finer granularity provided by the tree structure offers unique advantages, as each subtree can serve as another valid experience for decomposing subgoals, thus maximizing the utility of past experiences.

For \textbf{execution}, we aim to decompose goal accomplishment into multiple steps of tool invocations, and the ReACT reasoning chain precisely meets this requirement. For each goal $G$, its ReACT execution trajectory is represented by
$T_{G} = (s_1, s_2, \ldots, s_n)$,
where $T_{G}$ represents the trajectory with $n$ steps of tool invocations for the goal $G$. Each step $s_i$ {\small$(1 \leq i \leq n)$} may encapsulate meta-information such as the thoughts, tool name, tool inputs, and response from the tool, etc.
In this way, the whole execution trajectory, along with all its meta-information, serves as a holistic execution experience, guiding the agent to complete a goal effectively and efficiently. The ReACT structure refines the steps to complete a task, enabling greater flexibility in modifying, consolidating, and storing the experiences of achieving a specific goal.


Existing agent frameworks usually mix the planning and execution into one process. However, {XAgent} \citep{xagent2023} stands out by differentiating these two facets, achieved through the deployment of two distinct agent experts, each responsible for task planning and execution, respectively.
The planning system in XAgent is structured as a tree, with all its leaf node subgoals executed through a ReACT trajectory ($G_x$ is a leaf node subgoal \textit{is equivalent to} $T_{G_x}$ exists).
These traits all conform with the ideal agent system that maximizes the flexibility and utility of past experiences. In the following, we will mainly introduce our ICE strategy through the XAgent framework. Nevertheless, the core ideas behind our strategies could be generalized to other complex agent designs.




\paragraph{Problem Formulation.} Continuing with the symbols we defined above, we formalize the research question as follows:
For each goal $G^i$ in a multitude of past user goals
$\mathbf{G} = \{ G^1, \ldots, G^m \}$ {\small$(1 \leq i \leq m)$},
its past experiences involve the plan tree with $G^i$ as the root (planning experience) and a set of execution trajectories
$\mathbf{T} = \{ T_{G^{i}_{x}} \ | \ x = x_1\mhyphen x_2\mhyphen \dots \}$ (execution experiences).
Our goal is to apply all these past experiences from $\mathbf{G}$ to improve the effectiveness and efficiency of a new user goal $G^{\texttt{new}} \notin \mathbf{G}$.


%% file: sections/3_method.tex
\section{ICE Self-Evolution Strategy}

In this section, we present an in-depth explanation of the ICE self-evolution strategy for planning and execution. We divide the ICE strategy into \textbf{Planning self-evolution strategy} and \textbf{Execution self-evolution strategy}, each composed of three stages corresponding to \textsc{Investigate}, \textsc{Consolidate}, and \textsc{Exploit}.




\begin{figure*}[!t]
    \centering
    \subfigure{\includegraphics[width=0.85\textwidth]{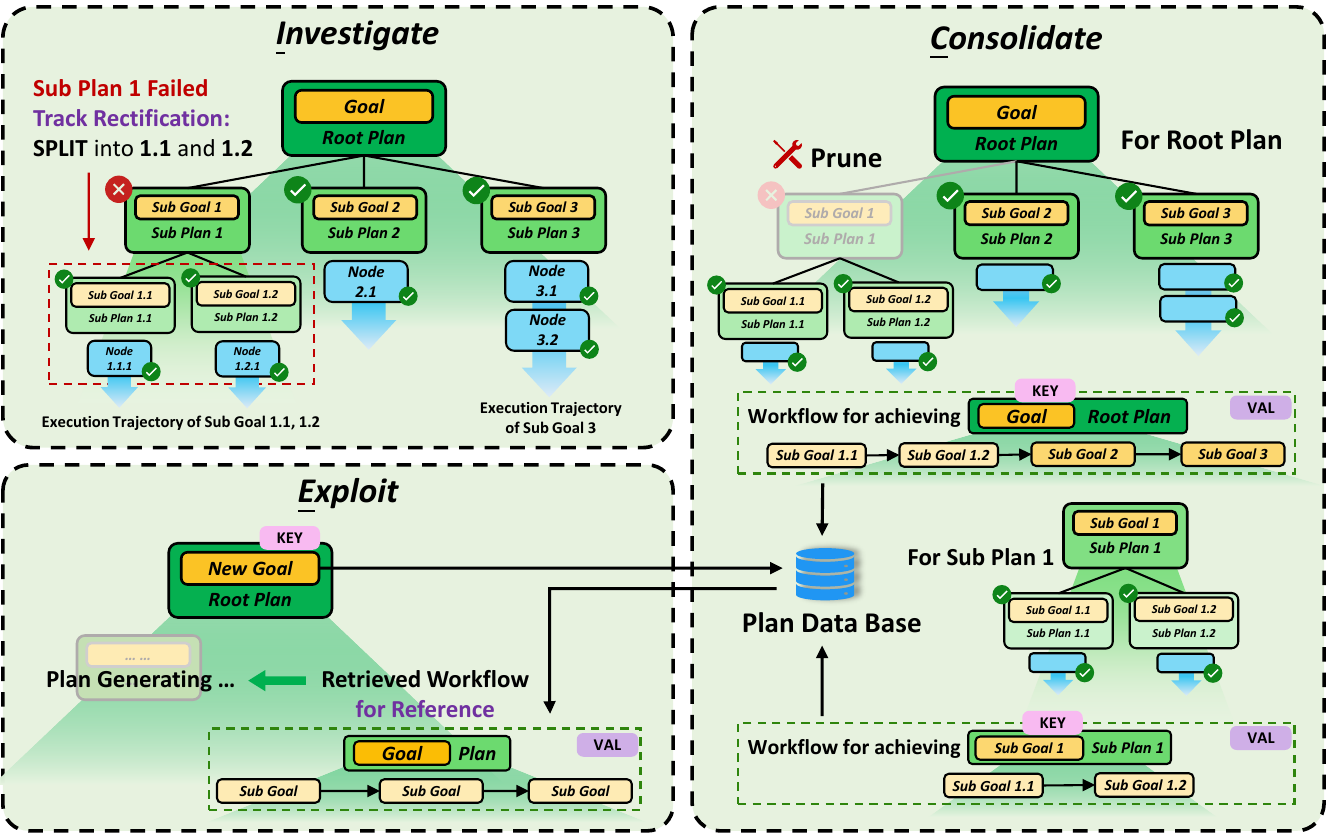}}
    \caption{
    \small{The detailed illustration of Planning ICE. Planning ICE dynamically tracks the plan of the agent system during \textsc{Investigate} stage, prunes and linearizes the plan into a workflow during \textsc{Consolidate} stage, and re-utilizes the workflow as a reference for new goals during \textsc{Exploit} stage.}
    }
    \label{fig:plan_ice}
\end{figure*}

\subsection{Planning ICE}

The ICE of planning involves dynamic tracking the plan during current task execution, pruning the plan as linearized workflows, and retrieving the workflows from memory as references for new goal decomposition.



\textbf{Plan-Investigate} 
The investigation of the plan aims to monitor the entire planning process and each goal's level of completion, facilitating the subsequent identification of valuable planning experiences.
This entails tracking: i) the initial plan generation (decomposition of the ultimate goal $G$ into subgoals), ii) any plan rectifications (e.g., splitting subgoal $G_i$ into $G_{i\mhyphen 1}$ and $G_{i\mhyphen 2}$, adding subgoal $G_{i+1}$ after $G_{i}$, etc.), and iii) the status of each goal or subgoal in the plan. Specifically, a goal or subgoal's status is assessed based on whether the corresponding milestones are achieved by its corresponding execution trajectory.

In \Cref{fig:plan_ice}, we show the \textsc{Investigate} stage tracks the split of a subgoal (in the red box) after its failure. The expected output is a finalized plan tree detailing all goals, subgoals, and their respective final statuses.



\textbf{Plan-Consolidate}
The consolidation of the plan aims to concatenate the successful goals and simplify the complex tree structure into a linear one easy for the agent to learn and refer to in the future.
This entails i) pruning all the failed goals or subgoals in the plan and ii) transforming the successfully achieved goals into a linear structure.

Specifically, we first gather a set of goals and subgoals whose final status is successful to form $\mathbf{G}^s$. For each $G_{x} \in \mathbf{G}^s$ {\small$(x = x_1\mhyphen x_2\mhyphen \dots)$}, if it is not a leaf node in the plan tree, we perform the following:
First, for the subtree extended from $G_{x}$, we prune all the subgoals in this subtree whose final status is failure. Next, we sequentially gather all the subgoals that are leaf nodes in this subtree, forming 
\begin{equation}
W_{G_x} = \{ G_{\texttt{leaf}} \ | \ \exists \ T_{G_{\texttt{leaf}}}, \ {\small \texttt{leaf} = x\mhyphen y_1\mhyphen y_2\mhyphen \dots} \},
\end{equation}
where, $G_{\texttt{leaf}}$ represents the leaf node under subgoal $G_x$. We define $W_{G_x}$ as the consolidated \textbf{workflow} of a successfully achieved goal $G_x$. For all the goals or subgoals in $\mathbf{G}^s$, we could thus derive a set of workflows $\mathbf{W} = \{ W_{G_x} \ | \ G_x \in \mathbf{G}^s \}$. All the consolidated workflows in $\mathbf{W}$ will then be stored in the agent system's memory, with the description of $G_x$ as key and $W_{G_x}$ as value.

In \Cref{fig:plan_ice}, we present in \textsc{Consolidate} stage the pruning of the failed subgoal and storage of two linearized workflows respectively corresponding to the root goal and a subgoal. The construction of workflows significantly lowers the learning barrier for future reference, and unifies the format of diverse planning experiences.



\textbf{Plan-Exploit} 
The exploitation of the plan aims to re-utilize the consolidated workflows as in-context references to improve the effectiveness of plan-making for the new user task goal.
This entails retrieving plans of similar goals both during \emph{initial plan generation} and \emph{plan rectifications}.

Specifically, during the initial plan generation of a new task goal $G^{\texttt{new}}$, we first retrieve workflow from the memory with a similar goal description. During the plan rectification after any subgoal fails, we retrieve the workflows that respectively align with i) the description of this failed subgoal $G^{\texttt{new}}_{x}$ {\small $(x = x_1\mhyphen \dots x_{k})$} and ii) description of its parent $G^{\texttt{new}}_{y}$ {\small $(y = x_1\mhyphen \dots x_{k-1})$} as a higher-level goal. Leveraging this prior knowledge, we guide the agent system to either i) find improved methods to achieve $G^{\texttt{new}}_{x}$ or ii) bypass $G^{\texttt{new}}_{x}$ to directly fulfill $G^{\texttt{new}}_{y}$. All the retrievals are based on the cosine similarity of the embeddings of the goals' descriptions.

In \Cref{fig:plan_ice}, we reveal how to take reference from the retrieved workflow during initial plan generation in \textsc{Exploit} stage. This ensures an effective and efficient planning process, aiding in the successful execution of subgoals later during self-evolution.

\begin{figure*}[!t]
    \centering
    \subfigure{\includegraphics[width=0.85\textwidth]{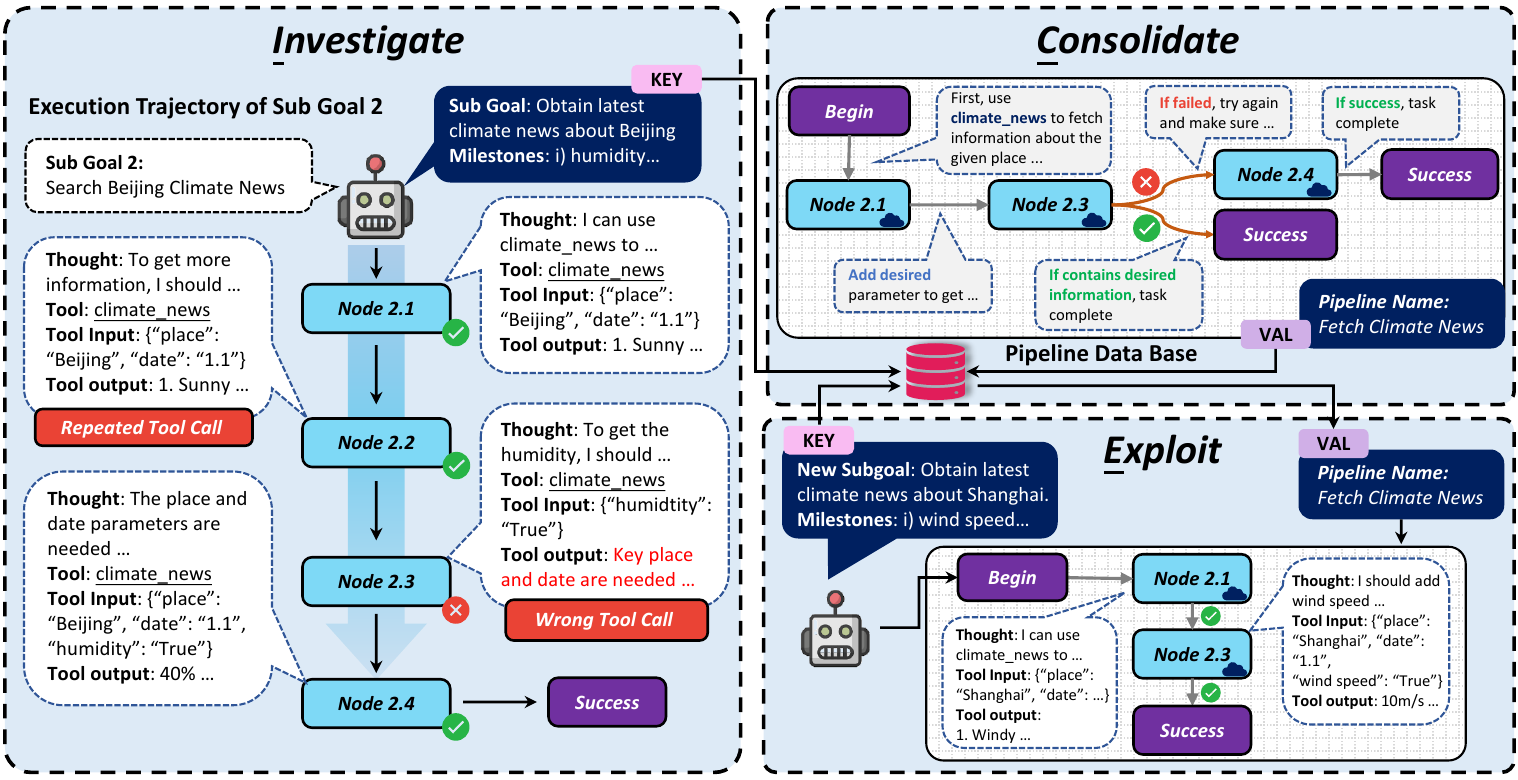}}
    \caption{
    \small{The detailed illustration of Execution ICE. Execution ICE identifies successful execution trajectories of the agent system during \textsc{Investigate} stage, extracts potential useful pipelines during \textsc{Consolidate} stage, and re-utilizes relevant pipelines for new goals during \textsc{Exploit} stage.}
    }
    \label{fig:pipeline_ice}
\end{figure*}

\subsection{Execution ICE}

The ICE of execution aims at utilizing the experiences gained from history to future inter-task execution. 
Intuitively, compared to the retrieved-augmented generation method, directly re-utilizing the execution trajectories in a restricted decoding manner is more efficient and can mitigate the cognitive burden of the agent.
It involves recording the execution trajectories, consolidating the trajectories into pipelines, and the retrieval and execution of pipelines from memory for new task execution.




\textbf{Execution-Investigate}
The investigation of the execution trajectories aims to identify all the trajectories that successfully achieve their corresponding subgoals.
Specifically, for all the leaf node subgoals $G_x$ in the plan tree, we record $T_{G_x}$ if the final status of $G_x$ is a success. This is because completely failed trajectories are more difficult to be directly taken as references during future task execution.
Note that though we do not include failed trajectories, we still include errors that occurred during certain steps, as long as they do not cause the whole trajectory to fail.


In \Cref{fig:pipeline_ice}, we present an execution trajectory with repeated and wrong tool calls in certain steps, but we still record it during \textsc{Investigate} stage as it completes the subgoal climate research.

\textbf{Execution-Consolidate}
The consolidation of execution aims to fix all the necessary steps of tool invocations to achieve a specific goal, thus facilitating automatic execution. This entails i) pruning all the unnecessary tool invocation steps, ii) adding more complex logic (e.g. switching logic) to make the trajectory to success more robust, and iii) explicitly suggesting how to transit to the next tool invocation step (e.g. which tool to use, how to fill in parameters).

We note that for the execution trajectory $T_{G_x} = (s_1, s_2, \ldots, s_n)$ aiming at subgoal $G_x$, if the tool name and parameters of every $s_i$ are fixed, then $T_{G_x}$ can be executed in a finite automaton manner. Therefore, we transform $T_{G_x}$ into the format of an automaton represented by:
\begin{equation}
A_{G_x} = ( (Q, \Sigma, \delta) ),
\end{equation}
where $Q$ denotes a set of tool invocation nodes, $\Sigma$ denotes edges representing rules or suggestions under which transitions occur, and $\delta$ denotes the transition function defined as $\delta: Q \times \Sigma \rightarrow Q$. We define $A_{G_x}$ as a consolidated \textbf{pipeline} for $G_x$. Note that for each node in a pipeline, the next step of tool invocation is restricted by its out-degree. This facilitates the agent to make more informed choices efficiently, in contrast to predicting arbitrarily within an open-ended ReACT trajectory.

The transformation from $T_{G_x}$ to $A_{G_x}$ is automatically done by prompting \texttt{GPT-4} with demonstrations (detailed in \Cref{apdx:pipeline_consolidation_examples}). All strategies, including filtering repeated or useless nodes and adding switching logic, are explicitly instructed and demonstrated in examples. The rules and suggestions on how to transit from one node to another are also prompted. Finally, we store $A_{G_x}$ into the agent system's memory with $G_x$ and its corresponding milestones as the key.

In \Cref{fig:pipeline_ice}, we present a consolidated pipeline with a switching structure. The previous repeated tool invocation (Node 2.3) is pruned, and the rule for the switching logic is explicitly added to avoid the irreversible failure brought by the wrong tool invocation (Node 2.3). Through consolidation into pipelines, the ReACT trajectories become more efficient and robust, while concurrently allowing for automatic execution.

\textbf{Execution-Exploit}
The exploitation of execution aims to directly apply the consolidated pipelines for use when handling new goals during future task execution. This entails retrieving pipelines according to the similarity of goals and executing the top-one pipeline.

Specifically, for a new subgoal $G^{\texttt{new}}_{x}$ in a new user task, we retrieve the pipeline with the most similar goal description and milestones.
Note that there's a threshold of similarity in retrieval, and the agent will fall back to ReACT execution if the similarity is not enough. If a pipeline is successfully retrieved, then the automaton execution process will start. During execution, the agent will either i) complete the tool parameters if there is only one outgoing edge, or ii) choose from multiple outgoing edges, following the suggestions of the outgoing edge(s) to move to the next node.



In \Cref{fig:pipeline_ice}, we show a successful exploitation of the stored pipeline to execute a similar subgoal about climate news searching. Flexible uses of pipelines can serve as a substitute for executing ReACT trajectories, thus effectively reducing the model API calls and making the whole execution more time-efficient and cost-efficient.

%% file: sections/4_experiment.tex
\section{Experiments}

To validate the effectiveness of \textsc{Investigate-Consolidate-Exploit} strategy, we conduct experiments on the XAgent framework \citep{xagent2023}.
Following the settings in ToolLLM~\citep{qin2023toolllm}, the agent system has access to high-quality RapidAPIs and external tools. We implement two ICE strategies based on this framework.

\subsection{Experimental Settings}

\paragraph{Data} As previous agent benchmarks do not involve complex tool invocation scenarios beneficial for intricate learning experiences, we manually create 40 tasks that encourage the invocation of diverse tools but are still within the agent's capabilities. These encompass a wide range of scenarios including trip planning, data analysis, review writing, etc. We randomly divide the tasks into 20 data points as the \emph{training set} and the other 20 data points as the \emph{testing set}. The training set is used for the \textsc{Investigate} and \textsc{Consolidate} stages, where the agent dynamically tracks the plans, identifies execution trajectories, and further consolidates them into workflows or pipelines stored in the agent memory. With these past experiences, the agent system then performs the tasks in the testing set to validate our strategy.

\paragraph{Settings} After performing the \textsc{Investigate} and \textsc{Consolidate} stages with the workflows and pipelines stored in agent memory, we first test on data points within the training set to ensure the feasibility of ICE in \emph{same task} self-evolution. Then, we generalize these prior experiences as resources for the self-evolution on \emph{tasks of similar distribution} in the testing set.
We also apply the Pinecone database as the agent's memory (detailed in \Cref{apdx:setting_details}).

\paragraph{Baseline} We apply \texttt{GPT-4} as the backbone model for the XAgent framework. We also involve \texttt{GPT-4} for the consolidation of execution trajectories into workflows as introduced previously. We compare against several baselines: i) Standard XAgent (No ICE applied, \texttt{GPT-4} or \texttt{GPT-3.5} based), ii) Only Planning ICE (\texttt{GPT-4} based), iii) Only Execution ICE (\texttt{GPT-4} based), iv) Planning and Execution ICE (\texttt{GPT-3.5} applied during \textsc{Exploit} stage).

\paragraph{Metrics} We objectively measure task performance through several dimensions. Please refer to \Cref{apdx:setting_details} for a more detailed definition.
\begin{itemize}[topsep=0pt, partopsep=1pt, leftmargin=12pt, itemsep=0pt]
    \item \textbf{API Calls}: We count the total model API calls and the number of API calls aiming at tool invocation. API calls explicitly reflect the cost-efficiency and implicitly reflect the time-efficiency.
    \item \textbf{Completion Rate}: We measure the passing rate of all the subgoals after executing the trajectory, which reflects the effectiveness in comprehensively achieving the user task goals.
    \item \textbf{Rectification Times}: We record the times of plan rectification, which reflects the effectiveness of plan-making.
    \item \textbf{Re-utilization Rate}: We record the re-utilization rate of the consolidated pipelines, which reflects the effectiveness of pipeline exploitation.
\end{itemize}

\begin{table*}[!t]
\begin{center}
\small
\tabcolsep=0.007\linewidth
\begin{tabular}{ccccccc}
\toprule
\textbf{\makecell{ICE Strategy}} & \textbf{Model} & \textbf{\makecell{API Calls\\(All)}} & \textbf{\makecell{API Calls\\(Tools)}} & \textbf{\makecell{Completion Rate\\(Subtasks, \%)}} & \textbf{\makecell{Rectifications\\Times}} & \textbf{\makecell{Re-utilization\\Rate}} \\
\midrule
\multirow{2}{*}{\textbf{\makecell{Standard (w/o ICE)}}} & \texttt{GPT-4}
& 3025 & 807 & 82.18 & 45 & - \\
& \texttt{GPT-3.5} & 4535 & 901 & 37.21 & 275 & - \\
\midrule
\textbf{\makecell{Planning ICE}} & \texttt{GPT-4}
& 2073 & 628 & 89.55 & 39 & - \\

\textbf{\makecell{Execution ICE}} & \texttt{GPT-4}
& 456 & 317 & 93.10 & - & 53.52 \\
\midrule
\multirow{2}{*}{\textbf{\makecell{Planning + Execution}}}
& \texttt{GPT-4} & 495 & 313 & 90.32 & 6 & 47.89 \\
& \texttt{GPT-3.5} & 401 & 257 & 90.74 & 5 & 53.52 \\
\bottomrule
\end{tabular}
\end{center}
\caption{\small{ICE results on the 20 training set tasks to test same task evolution.}}
\label{tab:training_set_result}
\end{table*}

\begin{table*}[!t]
\begin{center}
\small
\tabcolsep=0.007\linewidth
\begin{tabular}{ccccccc}
\toprule
\textbf{\makecell{ICE Strategy}} & \textbf{Model} & \textbf{\makecell{API Calls\\(All)}} & \textbf{\makecell{API Calls\\(Tools)}} & \textbf{\makecell{Completion Rate\\(Subtasks, \%)}} & \textbf{\makecell{Rectifications\\Times}} & \textbf{\makecell{Re-utilization\\Rate}} \\
\midrule
\multirow{2}{*}{\textbf{\makecell{Standard (w/o ICE)}}} & \texttt{GPT-4}
& 2265 & 745 & 72.97 & 107 & - \\
& \texttt{GPT-3.5} & 4071 & 880 & 25.33 & 234 & - \\
\midrule
\textbf{\makecell{Planning ICE}} & \texttt{GPT-4}
& 1779 & 532 & 86.36 & 35 & - \\

\textbf{\makecell{Execution ICE}} & \texttt{GPT-4}
& 443 & 318 & 94.44 & - & 39.44 \\
\midrule
\multirow{2}{*}{\textbf{\makecell{Planning + Execution}}}
& \texttt{GPT-4} & 540 & 384 & 90.00 & 6 & 47.89 \\
& \texttt{GPT-3.5} & 610 & 258 & 86.96 & 6 & 35.21 \\
\bottomrule
\end{tabular}
\end{center}
\caption{\small{ICE results on the 20 testing set tasks to test self-evolution on tasks of similar distribution.}}
\label{tab:testing_set_result}
\end{table*}

\subsection{Results}
We present our quantitative experimental results in \Cref{tab:training_set_result} and \Cref{tab:testing_set_result}, dividing the results of training and testing sets according to different backbone models and ICE strategies.

\textbf{Efficiency of ICE in Reducing API Calls.} Our findings indicate that both the planning and execution ICE significantly reduce the model's API calls. The combination of both strategies can reduce total calls by 80\%, demonstrating the substantial cost and time efficiency of the ICE strategy. A comparative analysis further reveals that the primary source of this reduction is the execution ICE. This is because the execution trajectory itself takes up most of the model API calls originally in the XAgent framework, and the consolidated pipelines can now serve as a more efficient substitute.

\textbf{Effectiveness of ICE in Task Execution and Plan Formulation.} Strategies incorporating ICE generally yield a higher subtask completion rate than the standard XAgent. Additionally, the number of rectifications required for the plan is considerably reduced. These results suggest that ICE enhances the effectiveness of subtask execution and the rationality and effectiveness of the plans formulated.

\textbf{Applicability of Extracted Pipelines.} The re-utilization rate of the mined pipelines from the training set tasks to the testing set is approximately 50\%. This rate indicates that the consolidated pipelines resulting from the execution ICE strategy are generally applicable to unseen scenarios, demonstrating our strategy's robustness.

\textbf{\texttt{GPT-3.5} vs. \texttt{GPT-4} Performance.} We apply \texttt{GPT-3.5} during the \textsc{Exploit} stage of ICE to raise efficiency during experience re-utilization. We discover using \texttt{GPT-3.5} as the backbone model during experience exploitation also exhibits a high completion rate and low rectification times, rivaling the performance of \texttt{GPT-4}. Moreover, it significantly outperforms the standard XAgent across all metrics. These findings suggest that the ICE strategy can significantly reduce the demand for the backbone model's capability when re-utilizing these consolidated experiences, thereby facilitating more cost-efficient agent deployment.

\subsection{Case Study}
To more closely examine the ICE strategy, we go over it with a case study, demonstrating three different scenarios.

\begin{figure*}[!t]
    \centering
    \subfigure{\includegraphics[width=\textwidth]{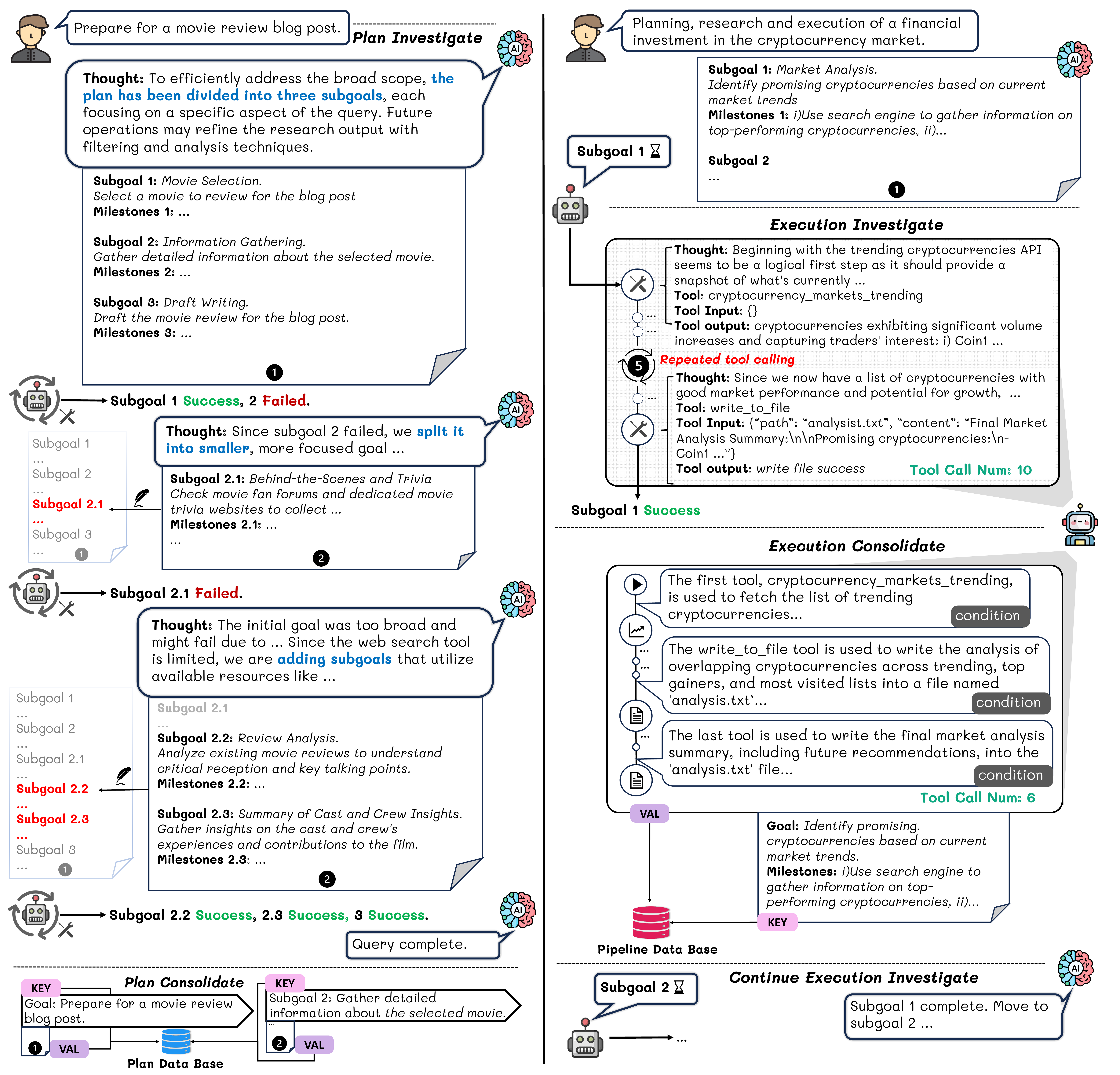}}
    \caption{
    \small{Case study on the \textsc{Investiagte} and \textsc{Consolidate} stages for the planning (left) and execution (right) self-evolution.}
    }
    \label{fig:investigate_consolidate_case}
\end{figure*}

\begin{figure*}[!t]
    \centering
    \subfigure{\includegraphics[width=\textwidth]{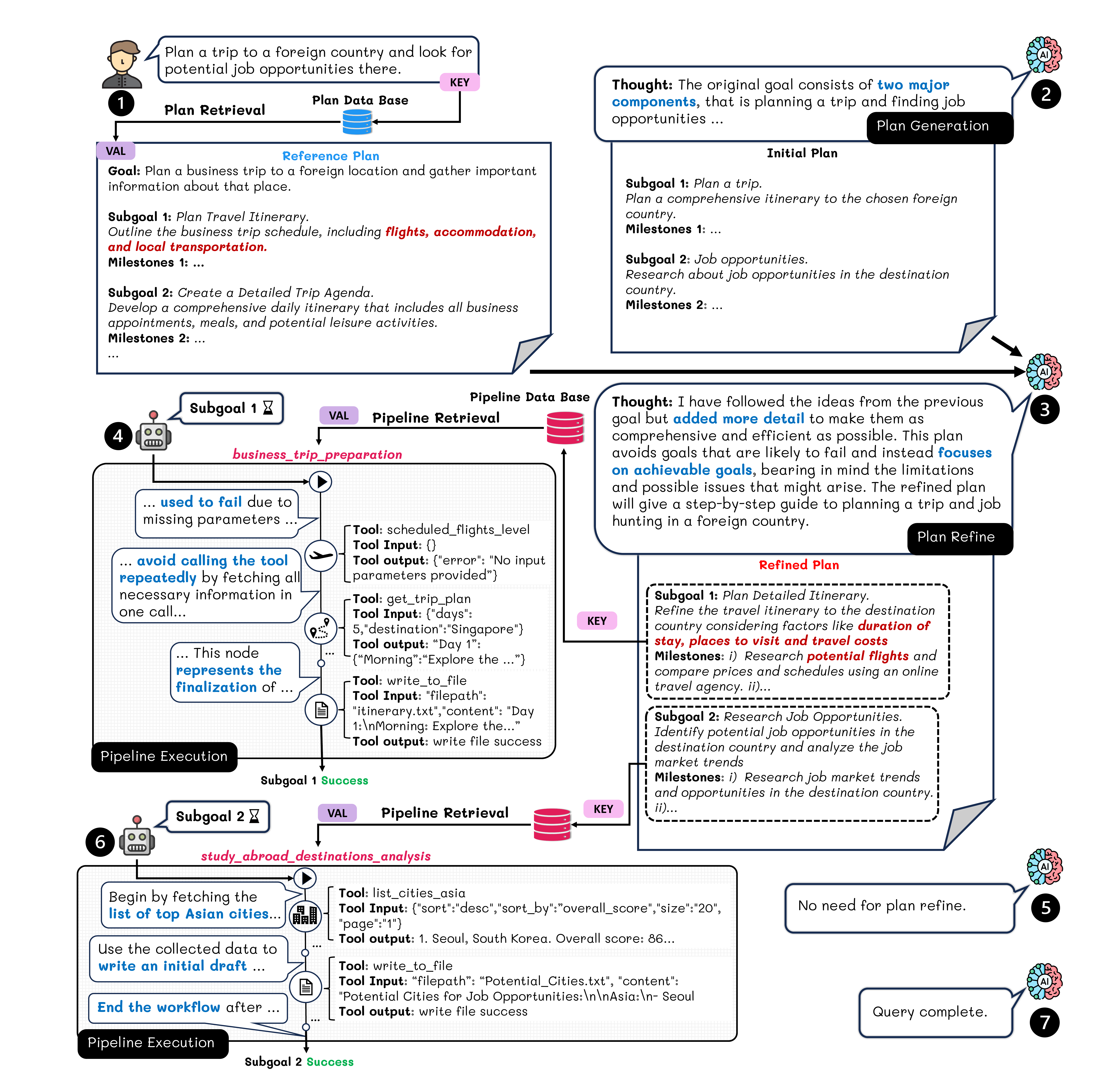}}
    \caption{
    \small{Case study on the \textsc{Exploit} stage for the planning (up) and execution (down) self-evolution.}
    }
    \label{fig:exploit_case}
\end{figure*}

\textbf{Planning Investigate and Consolidate}
As shown on the left of \Cref{fig:investigate_consolidate_case}, we construct a scenario about the preparation of a blog post. During this task execution, we dynamically track the status of each subgoal and any rectifications (e.g. the split of subgoal 2 into a smaller one 2.1, the addition of new subgoals 2.2 and 2.3). According to their final status, subgoals 2 and 2.1 are pruned from the tree during consolidation. Therefore, the final workflows are composed of subgoals \{1, 2.2, 2.3, 3\} for the ultimate user goal, and \{2.2, 2.3\} for subgoal 2. They would be stored in the plan database for future use.

\textbf{Execution Investigate and Consolidate}
As shown on the right of \Cref{fig:investigate_consolidate_case}, we construct another scenario about financial investment. For the first subgoal, we record the successful execution trajectory with repeated tool invocations. This trajectory is then consolidated into a static pipeline, where the repeated tool invocations are pruned for simplification, and the rules are explicitly added on edges to facilitate transitions between nodes. With the pipeline's goal and its corresponding milestones as the key, the pipeline is stored in the pipeline database for future use.

\textbf{Planning and Execution Exploit}
As shown in \Cref{fig:exploit_case}, the exploitation process involves retrieving workflows with a similar goal description from the plan database to serve as a reference for the initial plan generation of a new user task goal. Subsequently, each subgoal in the plan, along with its corresponding milestones, is utilized to retrieve a pipeline for automatic execution. This integrated approach ensures alignment with previously successful task goal decomposition structures, contributing to more effective and efficient execution of the current task.

\subsection{Ablation Study}

\begin{table}[!t]
\begin{center}
\small
\tabcolsep=0.02\linewidth
\begin{tabular}{cccc}
\toprule
\textbf{\makecell{Tasks for Experience Storage}} & \textbf{0} & \textbf{10} & \textbf{20} \\
\midrule
\textbf{\makecell{API Calls (All)}}
& 2265 & 750 & 540 \\
\textbf{\makecell{API Calls (Tools)}}
& 745 & 408 & 384 \\
\textbf{\makecell{Completion Rate (Subtasks, \%)}}
& 72.97 & 87.50 & 90.00 \\
\textbf{\makecell{Rectification Times}}
& 107 & 6 & 6 \\
\textbf{\makecell{Re-utilization Rate}}
& - & 44.44 & 47.89 \\
\bottomrule
\end{tabular}
\end{center}
\caption{\small{Results of ablation study on the number of tasks used for the storage of experiences.}}
\label{tab:ablation_result}
\end{table}

\textbf{Settings} We conduct an ablation study on the number of training tasks used for experience storage. The Standard XAgent setting represents zero prior experience, while previous experiments entail 20 tasks to store experiences. In this study, We randomly choose 10 tasks from the training set to store experiences, which are then exploited by the same 20 tasks in the testing set. We apply \texttt{GPT-4} as the base model for XAgent's execution.

\textbf{Results} In \Cref{tab:ablation_result}, we show as the number of tasks used for experience storage increases, the subtask completion rate gradually increases, while API calls significantly decrease. At the same time, more stored experiences could facilitate the pipeline's re-utilization. We discuss ICE's scaling effect in \Cref{sec:discussion} in detail.

%% file: sections/5_discussion.tex
\section{Discussions}
\label{sec:discussion}
The ICE strategies we proposed for self-evolution inspire many future research directions and open questions that we will discuss here.

\paragraph{Self-Evolution through Failed Experiences.} Our ICE strategy primarily consolidates successful experiences. However, during the \textsc{Investigate} stage, errors are inevitable and the information gathered from these failed experiences is also valuable. Successful experiences provide a straightforward path for the agent system to emulate, while failed experiences offer lessons on what to avoid. These failed experiences, however, are challenging to directly apply as they require the system's introspective capability. Future research could address this potential challenge by transforming failed experiences into explicit suggestions or rules, in order to avoid setting unattainable goals in the plan or invoking tools that are likely to fail during the execution.


\paragraph{Experience for In-context Learning or Direct Application.} During the \textsc{Exploit} stage, the workflows are used for in-context reference, and the pipelines are directly applied for re-utilization. This approach acknowledges that the user's \emph{high-level} goal is often not closely similar, making it challenging to directly apply the consolidated workflows to perform decomposition.
In contrast, the consolidated pipeline aims to execute a \emph{low-level} subgoal, which is more likely to be directly re-utilized under a similar \emph{high-level} goal. Both methods have strengths and weaknesses: i) In-context reference is more flexible but requires the model's attention to detail for successful adaptation. ii) Direct application is straightforward and more efficient in terms of time and cost, but the consolidated pipeline's goal may still not align perfectly with the current subgoal, making the evaluation of its effectiveness complex. Future research should focus on designing more robust methods to better balance these two approaches, as they constitute the core of self-evolution.



\paragraph{Scaling and Grokking of Agent Self-Evolution.} Our experiments have conclusively demonstrated the practicality of transforming prior experiences to facilitate new tasks. As larger memory volume allows for more precise and relevant retrieval, the accumulation of stored experiences can lead to \emph{scaling} and \emph{grokking} effects: the agent’s execution time and cost can be further reduced, while the tasks to which these experiences could generalize become more complex and diverse.
With the whole agent community involved, the collection of past experiences becomes more flexible and easier, as anyone can contribute and share their agent's task execution records for learning.
This makes our ICE strategy highly scalable and increasingly beneficial as more past experiences are accumulated.

%% file: sections/2_related_work.tex
\section{Related Work}

\paragraph{LLM-driven AI Agent.} Recent large language models (LLMs) have continuously demonstrated emerging intelligence \citep{wei2022emergent} in generating high-quality texts and codes \citep{zeng2022glm, chowdhery2022palm, openai2022chatgpt, openai2023gpt4, touvron2023llama}, performing robust reasoning \citep{wei2023chain, gao2023pal, yao2023tree, qian2023toolink}, and leveraging tools \citep{schick2023toolformer, qin2023tool, patil2023gorilla, qin2023toolllm}. These abilities enable LLMs to actively interact with the environment as agents, making plans and grounding actions while processing feedback \citep{xi2023rise, wang2023survey, yao2022react, ye2023large, ye2023proagent}. Current research into LLM agents involves the construction of robust agent frameworks \citep{qian2023creator, cai2023large, gur2023realworld}, exploration of multi-agent behaviors \citep{park2023generative, chan2023chateval, chen2023agentverse, li2023camel, qian2023communicative}, and benchmarks for agent evaluations \citep{zhou2023webarena, liu2023agentbench}. Apart from these directions, we investigate the capability of agent self-evolution through ICE.

\paragraph{Memory-Based LLM Enhancement.}
The implementation of the ICE strategy requires the agent's memory capacity. Early studies including memory-augmented networks \citep{meng2018dialogue, graves2014neural} and their computational universality \citep{schuurmans2023memory} utilized the memory matrix for the model's long-term information storage. Recent works focus on other memorizing mechanisms, such as MemoryBank inspired by Ebbinghaus’ forgetting curve theory \citep{zhong2023memorybank} and Think-in-Memory to imitate human-like recalling and post-thinking ability \citep{liu2023thinkinmemory}. Other external modules are also incorporated into the agent's architecture \citep{sumers2023cognitive}, enhancing LLM through external memory retrieval \citep{izacard2022atlas}, disentanglement of memory and knowledge \citep{wang2023augmenting}, and database as symbolic memory \citep{hu2023chatdb}. ICE utilizes external memory to store consolidated plans and pipelines, enabling LLM self-evolution through past experiences.

\paragraph{LLM Self-Improvement.} The concept of self-improvement is implicitly expressed in some current works, but not investigated in depth or uniquely as a strategy for agent designs. For most studies, this is demonstrated as the agent's iterative adaptation through task-execution \citep{madaan2023selfrefine, sun2023adaplanner}, code-execution \citep{qian2023creator}, or physical simulation \citep{song2023llmplanner, wang2023describe} feedback. Other self-evolution strategies take the form of prompt adaptation and optimization \citep{zhou2022large, yang2023large, pryzant2023automatic, hsieh2023automatic}, continuous improvement through error-identification and self-reflection \citep{shinn2023reflexion, miao2023selfcheck, qian2023creator}, and retrieval from database as short or long-term memory \citep{zhao2023expel, hu2023chatdb, liu2023thinkinmemory, zhong2023memorybank}. These works primarily emphasize the iterative refinement of a single task within a loop-structured LLM-based framework. In contrast, ICE confronts the challenges associated with inter-task agent self-evolution, offering strategies for the effective exploitation of past experiences.

%% file: sections/6_conclusion.tex
\section{Conclusion}
In this work, we introduce \textsc{Investigate-Consolidate-Exploit} (ICE), a novel strategy to facilitate the agent's inter-task self-evolution. ICE identifies the plan and execution trajectories as valuable past experiences for reference and re-utilization, thereby promoting the agent's continuous improvement. Through experiments on XAgent, we show the ICE strategy can reduce model API calls by up to 80\% and significantly lower requirements for model capabilities, thereby enhancing the deployment of agent systems in terms of time efficiency, cost efficiency, and overall task execution effectiveness.
Our work contributes to the burgeoning field of research focused on intelligent AI agents, highlighting the potential for complex agents to continuously learn from past experiences and adapt to diverse scenarios. We hope that our research and discussions will inspire a new paradigm in agent design, ultimately contributing to the development of a more robust ecosystem for AI agents.

%% file: sections/Appendix.tex
\clearpage
\appendix

\section*{Appendix}

\section{Demonstrations for Pipeline Consolidation}
\label{apdx:pipeline_consolidation_examples}
System prompt and in-context examples used in pipeline consolidation:

\noindent\makebox[\linewidth]{\rule{\linewidth}{0.4pt}}
\textit{System Prompt}\\
\begin{lstlisting}[basicstyle=\ttfamily, breaklines=true]
You are an experienced pipeline extractor who can extract rules and experiences given an execution trajectory.
You are given an execution trajectory with tool calls, which contain the tool name and tool input arguments. You need to generate some information describing what nodes and edges this pipeline contains:
1. some natural language comments and conditions explaining how the current tool call moves to the next tool call.
2. edges between tool calls
3. nodes for every tool call

Here are two examples:
{examples}

Note that:
- If one tool call appears more than once in the tool records, try to i) filter them and leave only one tool call node if those tool calls are useless repeated trials, ii) add switch logic as the example does, which means there are multiple out edges from the tool node.
- Always add the start node and end node in the nodes, and start edge and end edge in the edges.
- Try to simplify the pipeline. Avoid including the wrong tool call trials in the nodes and edges, instead add comments to the edges to state what should be noticed to avoid error happening or add error handle logic such as switch logic.
- Do not miss any properties in nodes and edges. Node name, tool name, and node type in nodes. Edge name, edge type, from node to node, and comments in edges.
\end{lstlisting}

\noindent\makebox[\linewidth]{\rule{\linewidth}{0.4pt}}
\textit{In-context Examples}
\begin{lstlisting}[basicstyle=\ttfamily, breaklines=true]
Example 1:
Query: Fetch the information of a product with sku W003247135 and W003247136.

Execution Trajectory:

Tool Name: RapidAPIEnv_rapi_wayfair_products _detail
Tool Arguments: {"sku": "W003247135"}
Tool Output: "response1"

Tool Name: RapidAPIEnv_rapi_wayfair_reviews _list
Tool Arguments: {"sku": "W003247135"}
Tool Output: "response2"

Tool Name: RapidAPIEnv_rapi_wayfair_products _detail
Tool Arguments: {"sku": "W003247136"}
Tool Output: "response3"

Tool Name: FileSystemEnv_write_to_file
Tool Arguments: {"filepath": "blog_post_material.txt", "content": "response1 + response2"}
Tool Output: "response1 + response2 + response"
\end{lstlisting}
\begin{lstlisting}[basicstyle=\ttfamily, breaklines=true]
Pipeline:
"pipeline_name": "product review fetch and write",
"pipeline_purpose": "Fetch overview information and details information of a given product.",
"nodes": [
{
    "node_name": "start",
    "tool_name": "Start",
    "node_type": "Start"
},
{
    "node_name": "end",
    "tool_name": "End",
     "node_type": "End"
},
{
    "node_name": "product_detail_1",
    "tool_name": "RapidAPIEnv_rapi_wayfair _products_detail",
    "node_type": "ToolServer"
},
{
    "node_name": "review_list",
    "tool_name": "RapidAPIEnv_rapi_wayfair _reviews_list",
    "node_type": "ToolServer"
},
{
    "node_name": "product_detail_2",
    "tool_name": "RapidAPIEnv_rapi_wayfair _products_detail",
    "node_type": "ToolServer"
},
{
    "node_name": "write_file",
    "tool_name": "FileSystemEnv_write_to_file",
    "node_type": "ToolServer"
}
],
"edges": [
{
    "edge_name": "start_product_detail",
    "edge_type": "data",
    "from_node": "start",
    "to_node": "product_detail_1",
    "comments": [
        "The first tool, RapidAPIEnv_rapi_wayfair _products_detail, is used to fetch the product details for the given SKU."
    ]
},
{
    "edge_name": "product_detail_review_list",
    "edge_type": "data",
    "from_node": "product_detail_1",
    "to_node": "review_list",
    "comments": [
        "The second tool, RapidAPIEnv_rapi_wayfair _reviews_list, is used to fetch the reviews for the same SKU."
    ]
},
{
    "edge_name": "review_list_product_detail_2",
    "edge_type": "data",
    "from_node": "review_list",
    "to_node": "product_detail_2",
    "comments": [
        "The third tool, RapidAPIEnv_rapi_wayfair _products_detail, is used to fetch the reviews for the SKU W003247136."
    ]
},
{
    "edge_name": "product_detail_2_write_file",
    "edge_type": "data",
    "from_node": "product_detail_2",
    "to_node": "write_file",
    "comments": [
        "The fourth tool, FileSystemEnv_write_to_file, is used to write the fetched product details and reviews into a file named 'blog_post_material.txt'."
    ]
},
{
    "edge_name": "end_pipeline",
    "edge_type": "data",
    "from_node": "write_file",
    "to_node": "end",
    "comments": []
}
]
\end{lstlisting}
\begin{lstlisting}[basicstyle=\ttfamily, breaklines=true]
Example 2:
Query: Fetch the information of a product with sku W003247135.

Execution Trajectory:

Tool Name: RapidAPIEnv_rapi_wayfair_products _detail
Tool Arguments: {"sku": "W003247135"}
Tool Output: "fail. Can not find product W003247135. Supported product: W003247136, W003247137, ..."

Tool Name: FileSystemEnv_write_to_file
Tool Arguments: {"filepath": "fail_reason_and_suggestions.txt", "content": "Reason: The current available product ids do not include sku W003247135.\nSuggestions: However, a similar product W003247136 can be obtained."}
Tool Output: "Reason: The current available product ids do not include sku W003247135.\nSuggestions: However, a similar product W003247136 can be obtained."

Tool Name: RapidAPIEnv_rapi_wayfair_products _detail
Tool Arguments: {"sku": "W003247136"}
Tool Output: "response 1"

Tool Name: RapidAPIEnv_rapi_wayfair_reviews_list
Tool Arguments: {"sku": "W003247136"}
Tool Output: "response 2"

Tool Name: FileSystemEnv_write_to_file
Tool Arguments: {"filepath": "blog_post_material.txt", "content": "response1 + response2"}
Tool Output: "response1 + response2"

Pipeline:
"pipeline_name": "product review fetch and write",
"pipeline_purpose": "Fetch overview information and details information of a given product.",
"nodes": [
{
    "node_name": "start",
    "tool_name": "Start",
    "node_type": "Start"
},
{
    "node_name": "end",
    "tool_name": "End",
    "node_type": "End"
},
{
    "node_name": "product_detail",
    "tool_name": "RapidAPIEnv_rapi_wayfair _products_detail",
    "node_type": "ToolServer"
},
{
    "node_name": "write_fail_reason_and_suggestions",
    "tool_name": "FileSystemEnv_write_to_file",
    "node_type": "ToolServer"
},
{
    "node_name": "product_detail_retry",
    "tool_name": "RapidAPIEnv_rapi_wayfair _products_detail",
    "node_type": "ToolServer"
},
{
    "node_name": "review_list",
    "tool_name": "RapidAPIEnv_rapi_wayfair _reviews_list",
    "node_type": "ToolServer"
},
{
    "node_name": "write_obtained_information",
    "tool_name": "FileSystemEnv_write_to_file",
    "node_type": "ToolServer"
}
],
"edges": [
{
    "edge_name": "start_product_detail",
    "edge_type": "data",
    "from_node": "start",
    "to_node": "product_detail",
    "comments": [
        "The first tool, RapidAPIEnv_rapi_wayfair _products_detail, is used to fetch the product details for the given SKU."
    ]
},
{
    "edge_name": "product_detail_write_fail _reason_and_suggestions",
    "edge_type": "data",
    "from_node": "product_detail",
    "to_node": "write_fail_reason_and_suggestions",
    "comments": [
        "Here is a switch logic: If the response from node product_detail is failed, which means the RapidAPIEnv_rapi_wayfair _products_detail tool do not support the product SKU given in the user query, FileSystemEnv_write_to_file, is used to write the failed reason and suggestions into a file named 'fail_reason_and_suggestions.txt'."
    ]
},
{
    "edge_name": "write_fail_reason_and_suggestions _product_detail_retry",
    "edge_type": "data",
    "from_node": "write_fail_reason_and_suggestions",
    "to_node": "product_detail_retry",
    "comments": [
        "Retry the RapidAPIEnv_rapi_wayfair _products_detail tool with suggestions written before."
    ]
},
{
    "edge_name": "product_detail_retry _review_list",
    "edge_type": "data",
    "from_node": "product_detail_retry",
    "to_node": "review_list",
    "comments": [
        "Use the response from node product_detail_retry to review_list, the RapidAPIEnv_rapi_wayfair _reviews_list tool, to fetch the reviews for the suggested SKU."
    ]
},
{
    "edge_name": "product_detail_review_list",
    "edge_type": "data",
    "from_node": "product_detail",
    "to_node": "review_list",
    "comments": [
        "product_detail node appears the second time here, so here is another possible option for the switch logic: If the response from node product_detail is successful, then RapidAPIEnv_rapi_wayfair _reviews_list, is used directly to fetch the reviews for the same SKU."
    ]
},
{
    "edge_name": "review_list_write_obtained _information",
    "edge_type": "data",
    "from_node": "review_list",
    "to_node": "write_obtained_information",
    "comments": [
        "The next tool, FileSystemEnv_write_to_file, is used to write the fetched product details and reviews into a file named 'blog_post_material.txt'."
    ]
},
{
    "edge_name": "end_pipeline",
    "edge_type": "data",
    "from_node": "write_obtained_information",
    "to_node": "end",
    "comments": []
}
]
\end{lstlisting}

\section{Experimental Setting Details}
\label{apdx:setting_details}
We apply the Pinecone vector database as the agent system's memory. For the consolidated workflows, we use the ultimate user's goal or the subgoals as the key and store the whole corresponding workflow. For the consolidated pipelines, we use the subgoal and its milestones as the key and store the pipeline with its name. The pipeline is consolidated into JSON format.
The embedding of the key is derived by calling OpenAI \texttt{text-embedding-ada-002}. The retrievals of workflows and pipelines are all based on the embedding's cosine similarity.

For the metrics we apply, the total model API calls refer to all the calls of the backbone model regarding planning and execution, while excluding the callings of \texttt{text-embedding-ada-002} for embeddings. The calls for tool invocation under the XAgent framework refer to how many times the \texttt{handle\_subtask} function is called during execution. This function is in charge of deciding how to handle the current subgoal and which tool to invocate, thus representing the real execution steps in the trajectory. The completion rate is measured by if the trajectory ends explicitly with \texttt{success} as the final status with all the milestones in the corresponding subgoal achieved.
